\documentclass[nonacm,sigplan]{acmart}

\usepackage{adjustbox}
\usepackage{multirow}
\usepackage{hyperref}

\newcommand{\sys}{LLM-PQ}
\newcommand{\sysplain}{LLM-PQ}
\newcommand{\basef}{PipeEdge}
\newcommand{\bases}{Uniform}

\newcommand{\basefr}{FlexGen}
\newcommand{\baseff}{FlexGen-int8}

\usepackage{mathtools}
\usepackage{algorithm, algorithmic}
\DeclarePairedDelimiter\ceil{\lceil}{\rceil}
\DeclarePairedDelimiter\floor{\lfloor}{\rfloor}

\newtheorem{theorem}{Theorem}

\usepackage{graphicx,subfigure}

\usepackage{listings}
\definecolor{dkgreen}{rgb}{0,0.6,0}
\definecolor{gray}{rgb}{0.5,0.5,0.5}
\definecolor{mauve}{rgb}{0.58,0,0.82}
\newcommand{\cwu}[1]{\textcolor{red}{[CWu: #1]}}

\author{Juntao Zhao}
\authornote{Work done as an intern at ByteDance Inc.}
\authornote{Corresponding author: \{juntaozh\}@connect.hku.hk}
\affiliation{
    \institution{University of Hong Kong}
    \country{Hong Kong}
}
\author{Borui Wan}
\affiliation{
    \institution{University of Hong Kong}
    \country{Hong Kong}
}

\author{Yanghua Peng}
\affiliation{
    \institution{ByteDance Inc.}
    \country{USA}
}

\author{Haibin Lin}
\affiliation{
    \institution{ByteDance Inc.}
    \country{USA}
}

\author{Chuan Wu}
\affiliation{
    \institution{University of Hong Kong}
    \country{Hong Kong}
}

\renewcommand{\shortauthors}{Zhao et al.}

\begin{document}

\title{\sysplain{}: Serving LLM on Heterogeneous Clusters with Phase-Aware Partition and Adaptive Quantization}


\begin{abstract}
Recent breakthroughs in Large-scale language models (LLMs) have demonstrated impressive performance on various tasks. The immense sizes of LLMs have led to very high resource demand and cost for running the models. Though the models are largely served using uniform high-caliber GPUs nowadays, utilizing a heterogeneous cluster with a mix of available high- and low-capacity GPUs can potentially substantially reduce the serving cost. There is a lack of designs to support efficient LLM serving using a heterogeneous cluster, while the current solutions focus on model partition and uniform compression among homogeneous devices.
This paper proposes \sysplain{}, a system that advocates adaptive model quantization and phase-aware partition to improve LLM serving efficiency on heterogeneous GPU clusters. We carefully decide on mixed-precision model quantization together with phase-aware model partition and micro-batch sizing in distributed LLM serving with an efficient algorithm, to greatly enhance inference throughput while fulfilling user-specified model quality targets.
Extensive experiments on production inference workloads in 11 different clusters demonstrate that \sysplain{} achieves up to 2.88$\times$ (2.26$\times$ on average) throughput improvement in inference, showing great advantages over state-of-the-art works. Source code available at https://github.com/tonyzhao-jt/LLM-PQ.
\end{abstract}

\maketitle 
\pagestyle{plain} 

\section{Introduction} 

Large-scale language models (LLMs) such as GPT3, LLaMA, OPT, and BLOOM~\cite{scao2022bloom, Zhang2022OPTOP, Touvron2023LLaMAOA} have exhibited unprecedented performance in pushing the envelope of various artificial intelligence (AI) tasks. The outstanding model performance 
is largely attributed to a very large model size ranging from a few hundred million to even half a trillion parameters. Training an LLM requires thousands of GPUs and millions of dollars \cite{gpt3}. Serving a trained LLM is also resource-demanding and cost-intensive, as an LLM cannot commonly be fit into a single GPU, therefore multiple GPUs are required for distributed inference. 


To cope with the massive size of LLMs, a number of approaches have been proposed to enable their efficient deployment in practice. 
DeepSpeed~\cite{Aminabadi2022DeepSpeedIE}, FasterTransformer and HuggingFace 
 Text Generation Inference (TGI)~\cite{huggingface_text_generation_inference} 
integrate existing model parallelism techniques, such as tensor-parallelism (TP) and pipeline parallelism (PP), with memory footprint reduction schemes, e.g., quantization or offloading, to lower the resource demands of model serving 
in a distributed manner. 
For memory footprint reduction schemes, quantization converts model weights into lower-precision formats (e.g., 8-bit), reducing memory consumption. Offloading methods~\cite{flexgen} leverage aggregate CPU and NVMe memory capacity to store weights or compute a portion of the GPU workload. However, the existing solutions are mainly designed for models serving on homogeneous clusters, limiting their performance in a heterogeneous cluster.

\begin{figure}[t]
    \centering
    \subfigure[GPU Portions]{
        \includegraphics[width=0.4\linewidth]{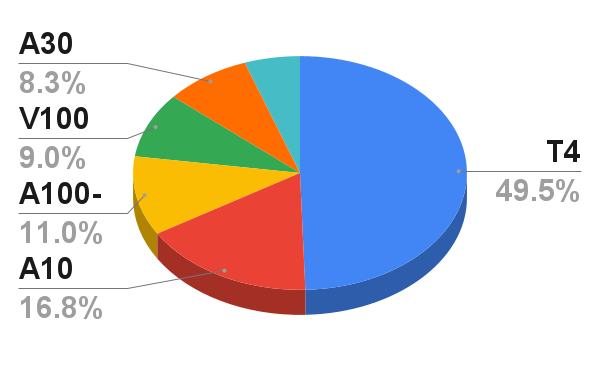}
    }
    \subfigure[Average utilization of different types of GPUs in one month] 
    {
        \includegraphics[width=0.5\linewidth]{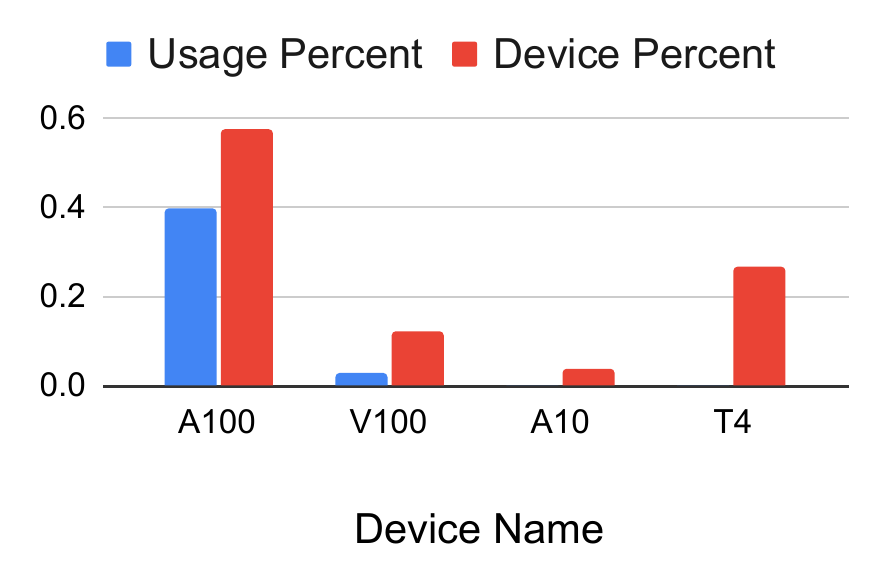}
    }
    \caption{GPU proportions and utilization rates in a real-world production AI cluster.}
    \label{fig:cluster_info}
\end{figure}

A practical AI cloud or machine learning (ML) cluster often contains heterogeneous devices, e.g., GPUs of different models purchased at different times. Utilization of different types of GPUs may differ substantially. Fig.~\ref{fig:cluster_info} shows the proportion of different GPUs in a production cluster, with fewer percentages of high-calibre GPUs (NVIDIA A100, V100) 
the majority being relatively low-calibre inference GPUs (such as T4). The utilization rate of other GPUs is much lower than that of A100, which are used intensively for both training and inference of large models nowadays for the best performance. \textbf{Efficiently exploiting available heterogeneous GPUs for LLM serving} is worthwhile to explore, to fully utilize available resources and substantially reduce the cost of provisioning LLM-enabled applications. 
The commonly adopted 
TP and PP paradigms 
partition model operations/layers evenly among the GPUs, which is not suitable for heterogeneous GPUs and results in either low utilization of high-capacity GPUs or out-of-memory (OOM) errors on low-memory GPUs. The limited studies of models serving on heterogeneous clusters~\cite{hu2021pipeline} focus on the partition of encoder-based transformer models. 
However, mainstream LLMs with decoder-only structures contain two phases during inference: prompt processing (prefill) and token generation (decode). While the former phase is similar to the inference of encoder-based transformers, the latter has a totally different pattern (see Sec.~\ref{sec:motive_llm}), making the previous partition solutions not suitable. Besides, the execution time required for each phase, depending on the prompt length and token generation number, varies significantly. What is worse, in a heterogeneous cluster, this difference can even be amplified, causing model partitioning that focuses on the time of the first phase instead of both being far from optimal. Therefore, phase-aware model partition schemes warrant investigation.   
Additionally, extra memory required for pre-and post-processing during LLM inference, such as text embedding for converting input tokens to word vectors, should also be considered, especially when utilizing low-calibre GPUs which have limited GPU memory. 


When the model is partitioned among heterogeneous GPUs, adopting a single quantization precision across all model layers in different types of GPUs is always suboptimal. 
uniform single-precision model quantization can select a precision, e.g., INT4, that is suitable for GPUs with lower memory to avoid OOM (Out Of Memory) problem, but causing a notable portion of memory waste for those with abundant GPU memory. 
 Adaptive mixed-precision quantization for LLM, which is not investigated in the literature~\cite{frantar2023gptq, xiao2023smoothquant}, is more desirable. By using higher precision for model weights on GPUs with more available memory instead of forcing them to use the same one in those low-calibre GPUs, adaptive mixed-precision quantization can not only avoid memory waste but promote the model quality as well. 

In this work, we propose a novel system, \sysplain{}, to enable efficient LLM generative serving on heterogeneous GPU clusters. Instead of emphasizing the enhancement of throughput faced with \textbf{infinite requests}, as commonly pursued in recent works like vLLM~\cite{vllm}. \sysplain{} directs its focus toward the efficient processing of a given workload, which is faced by the \textbf{offline task}. \sysplain{} advocates adaptive model quantization and phase-aware model partition, as well as efficient micro-batch scheduling for LLM pipeline serving. It jointly determines the quantization precisions, model layer partition, and hybrid micro-batch sizing strategies, given the LLM, available resources of the heterogeneous cluster, and user-specified model quality targets.
Our contributions in designing \sysplain{} can be summarized as follows: 


$\triangleright$ We provide a cost model that details the memory requirements of LLM serving under a mixed-precision quantization scheme. We learn a linear regression model to accurately predict the latency of mixed-precision LLM inference workloads with varying 
sequence lengths and batch sizes based on their phase-aware computational characteristics. 


$\triangleright$ 
We introduce adaptive mixed-precision into the search space of heterogeneous pipeline serving of LLM and provide a variance indicator the measure the layer sensitivity towards different level quantization. 
We develop an iterative algorithm that first explores possible GPU 
orderings and different (phase, micro-batch size) pairs in the pruned search space, 
and then solves an integer linear programming (ILP) problem to determine the best partition and quantization bitwidths. 

$\triangleright$ We have implemented a prototype of \sysplain{}, including the serving pipeline, a thread-safe micro-batch scheduler, and an 
on-the-fly quantized weight loader. 
 We extensively evaluate \sysplain{} 
 under various settings on 11 clusters composed of the most common GPU types (e.g., T4, P100, V100, A100, and A800). Experimental results demonstrate that our cost model incurs less than 6\% prediction errors and our LLM serving achieves up to 2.88$\times$ throughput improvement (2.26$\times$ on average) as compared to state-of-the-art approaches. 

%

\section{BackGround and Motivation}\label{sec:motive}
\subsection{Generative Inference of LLM}\label{sec:motive_llm}
\begin{figure}[t]
    \centering
    \includegraphics[width=0.98\linewidth]{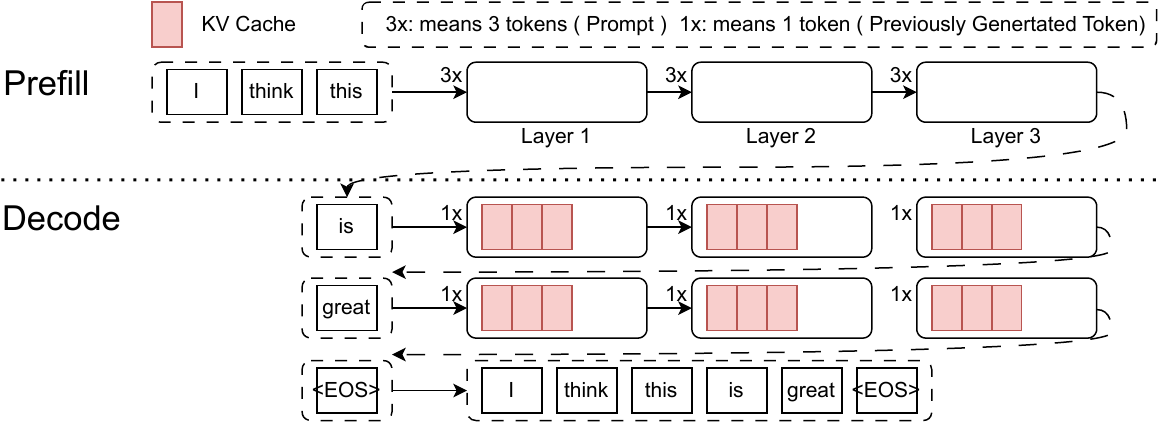}
    \caption{Two phases in LLM generative serving: (Top) \textbf{Prefill phase} takes the prompt sequence to generate the initial key-value pairs. (Bottom) \textbf{Decode phase} takes previously generated token \& stored KV pairs to generate the next token.} 
    \label{fig:two_stage_vis}
\end{figure}

\textbf{LLM} generally refers to a suite of decoder-only 
transformer models with large parameter sizes~\cite{Zhang2022OPTOP, scao2022bloom}. 
Unlike 
encoder-based transformers like ViT-Huge~\cite{vit} and Bert-Large~\cite{bert} 
that 
are sequence-to-sequence, 
LLMs generate tokens one by one in an inference process that comprises two phases~\cite{Zhang2022OPTOP, scao2022bloom}(Fig.~\ref{fig:two_stage_vis}): \textbf{prefill} and \textbf{decode}~\cite{flexgen}. In the prefill phase, the input prompt sequence produces key/value (KV) caches for each transformer layer, which is used in the attention mechanism as a context vector for later token generation. 
During the decode phase, stored KV pairs are updated as each subsequent token is generated one by one based on the preceding token; 
the token generation process continues until a stopping criterion is met, such as reaching the end of a sequence (EOS) or exceeding the maximum number of tokens allowed. During generative inference, each layer of the LLM undergoes a prefill phase followed by several passes in the decode phase (an example is given in Fig.~\ref{fig:two_stage_vis}).

The time taken by the prefill and decode phases varies to the prompt length. By sampling 10,000 conversations generated by chatGPT from the ShareGPT~\cite{ryokoai_sharegpt52k} dataset, 
we found that the prompt length varies substantially: 
$<128$ (14.20\%), $129$-$512$ (20.52\%), $513$-$1024$ (14.24\%), $1025$-$2048$ (14.53\%) and others (36.51\%).  
In the upper part of Fig.~\ref{fig:two_stage_cost}, 
we evaluate the time required to process a batch of 8 sequences and generate 32 tokens per sequence, with prompt lengths of 1024 and 128 on opt-13b and opt-30b models~\cite{Zhang2022OPTOP}, respectively. The prefill time increases with the prompt length (as it processes all the prompt tokens once) and is substantial ($\geq 36\%$) 
when the prompt is long. Unlike prefill time, the decode time is determined by the number of generated tokens. These characteristics make the inference pattern of LLM more complicated than encoder-based transformers.

\subsection{Heterogenous Model Parallelization}
\begin{figure}[t]
    \centering
    \includegraphics[width=\linewidth]{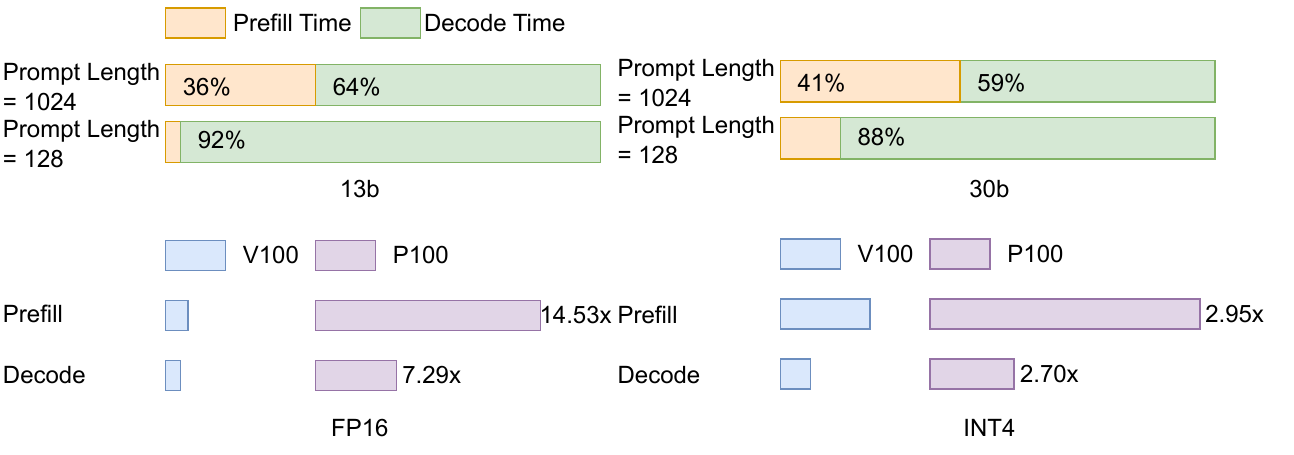}
    \caption{Phase time decomposition with different precisions. $\times$ indicates time on P100 compared to V100.}
    \label{fig:two_stage_cost}
\end{figure}

\noindent


Pipeline parallelism \cite{Aminabadi2022DeepSpeedIE, huang2019gpipe} has been widely adopted 
to distribute massive parameters of LLM across devices. 
The model is split into stages and micro-batches are processed over the stages in a pipelining manner. Each device executes a model stage, and data is passed between devices as it moves through the pipeline. Workload balance among stages is important as the throughput of pipeline serving is bounded by the execution time of the slowest stage. 

Deriving optimal partitions among heterogeneous devices 
is 
challenging, especially when considering the two-phase token generation. 
The lower part of Fig.~\ref{fig:two_stage_cost} gives 
the execution time of a single layer of the respective model 
with prompt length 512 and batch size 8. The execution time ratio when running the same phase on different devices varies substantially. 
For example, under FP16, the execution time of the layer in the prefill phase on P100 is 14.53$\times$ larger than that on V100, while the execution time ratio is 7.29$\times$ for the decode phase. Since the LLM inference time contains these two phases, 
pipeline stage partitioning 
should consider the execution time of both phases on each GPU. 
Existing solutions (e.g., PipeEdge~\cite{hu2021pipeline}) partition single-phase encoder-based models on heterogeneous devices, whose solutions cannot be directly extended to two phases of LLM serving. 

Furthermore, a complete language model includes an embedding layer, which is responsible for converting sentences into word vectors. In heterogeneous clusters, the embedding layer encounters more significant imbalance issues compared to homogeneous clusters due to the variety of the GPU's computing and memory capabilities.

\subsection{Online and Offline Serving Task}
There are two suites of the LLM inference workload. The \textbf{online task} handles infinite requests from runtime users, where the prompt length and token generation number are unpredictable. vLLM~\cite{vllm} introduces pageAttention to efficiently manage substantial and dynamically changing KV caches for each request. The \textbf{offline task} consists of predictable batch prompt processing tasks, where prompts are padded to a uniform length and the number of token generations is predetermined. FlexGen~\cite{flexgen} addresses the memory constraints in this scenario by employing multi-hierarchy offloading and zig-zag packing techniques.

\sysplain{} targets the \textbf{offline task} with the prior knowledge of the prompt length and token generation number.

\noindent
\textbf{Opportunity 1: Phase-Aware Model Partition on Heterogenous GPUs.}
By considering the inference time of both the prefill and decode phases, and also taking into account the resource consumption and computation of the embedding layer, we can obtain a more comprehensive understanding of the complex generation process of LLM and therefore a more accurate latency modeling for making pipeline stage partition decision. This approach ensures improved performance in heterogeneous pipeline serving.




\subsection{Quantization in LLM}

\textbf{Quantization} is a model compression technique that maps high-precision values, such as those stored in FP16, to their low-precision counterparts. For symmetric quantization, the input data or model weight distribution is evenly partitioned into a fixed number of bins. Each bin is rounded to an n-bit quantized value 
using $\hat{x} = [\frac{x-q_x}{s_x}]$, where $x$ is the original value in floating-point format, $q_x$ and $s_x$ are the zero-point and scaling factor, respectively, $[\cdot]$ is the rounding function and $\hat{x}$ is the resulting quantized value in lower-precision form. 
For each element $x \in$ vector $\mathbf{x}$, the scaling factor is derived 
as $s_x = \frac{\mathbf{x}_{max} - \mathbf{x}_{min}}{2^{b} - 1}$, where $\mathbf{x}_{max}$ and $\mathbf{x}_{min}$ are maximum and minimum values of the vector, 
and $b$ is the bitwidth. Dequantization is done with $\tilde{x} = s_{x}\hat{x}  + q_x$, where $\tilde{x}$ is the dequantized value in floating point. 

\noindent\textbf{LLM Quantization.} 
The weights of LLMs are typically stored in FP16/BF16. Due to the large size of LLMs, 
it is often necessary to further compress the model weights for inference serving, e.g., using 
INT8 quantization to reduce the weight storage by half. 
Existing LLM quantization approaches can be categorized into two: (1) 
W8A8 kernel-based quantization (e.g., 
SmoothQuant~\cite{xiao2023smoothquant} and ZeroQuant\cite{yao2022zeroquant}), which quantizes both activations and weights during serving; (2) weight-only quantization 
\cite{frantar2023gptq, frantar2023optq, lin2023awq}, which only quantizes model weights when loading the model into GPU memory. In this paper, we adopt decomposition kernel-based HuggingFace bitsandbytes~\cite{dettmers2022llmint8} to implement INT8 quantization. For precisions lower than 8 (e.g., 3 and 4), we use weight-only kernels provided by GPTQ~\cite{frantar2023gptq} following serving system setup of HuggingFace TGI~\cite{huggingface_text_generation_inference}, OpenLLM~\cite{Pham_OpenLLM_Operating_LLMs_2023}. 
However, Existing LLM quantization works 
uniformly quantize all model layers to the same bit by default (e.g., 3, 4, or 8~\cite{frantar2023gptq, dettmers2022llmint8}) 
which leads to 
underutilized memory on high-calibre GPUs or OOM problems on low-calibre GPUs in a heterogeneous cluster. This is because different types of GPUs are not allowed to choose their most suitable quantization precision to match their capacities. 

\begin{figure}[t]
    \centering
    \subfigure[BlOOM-3b PPL vs.~Bitwidth]{
        \includegraphics[width=0.47\linewidth]{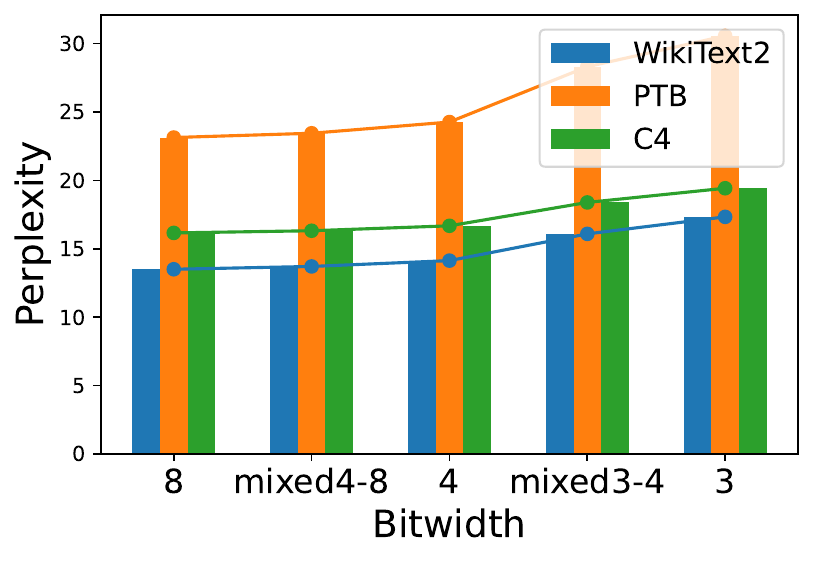}
    }
    \subfigure[OPT-1.3b Accuracy vs.~Bitwidth]{
        \includegraphics[width=0.47\linewidth]{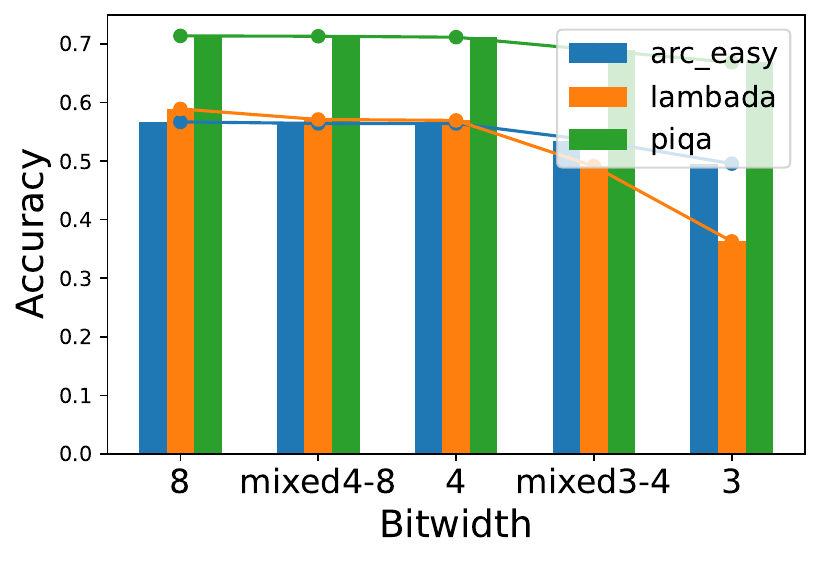}
    }
    \caption{BLOOM-3b (a) and OPT-1.3b (b) perplexity (PPL) \& accuracy under different quantization schemes. Smaller PPL means the model is more confident in its prediction.
    }
    \label{fig:vary_bit_acc}
\end{figure}

\noindent\textbf{Opporunity 2: Adaptive Quantization for Better Accuracy and Speed.}
We advocate adaptive quantization by choosing potentially different bits for model layers on different GPUs, to better utilize the available memory, as well as to improve model quality and computation speed as compared to uniform quantization. We illustrate the benefits of adaptive quantization as follows:

\noindent{\em 1. Adaptive quantization can lead to better model accuracy}. 
We run BLOOM-3b1~\cite{scao2022bloom} and OPT-1.3b~\cite{Zhang2022OPTOP} with different precision setups on A100
and evaluate the \textit{perplexity}~\cite{jelinek1977perplexity}, 
on three text datasets~\cite{wikitext2, ptb, c4}. We also measure the model \textit{accuracy} on popular zero-shot 
question-answering benchmarks LAMBADA~\cite{paperno-etal-2016-lambada}, ARC\cite{allenai:arc} and PIQA~\cite{Bisk2020};
We use calibration data from the C4 dataset 
to determine quantization statistics.  
In Fig.~\ref{fig:vary_bit_acc}, the `mixed4-8' case denotes that we uniformly randomly assign 4 or 8 bits to each model layer, while `mixed3-4' is to uniformly randomly assign bitwidth 3 or 4 to each layer. 
Mixed-precision quantization leads to better model performance than uniformly using the lower bit.

\begin{figure}[t]
    \centering
    \includegraphics[width=0.84\linewidth]{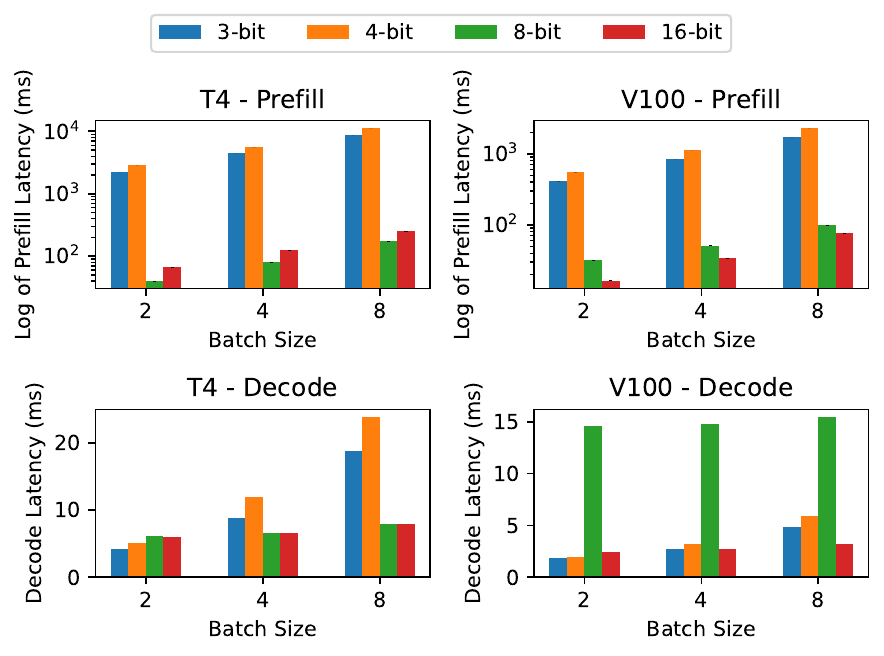}
    \caption{
    Execution time of prefill and decode phases under different precisions and batch sizes.} 
    \label{fig:motive_perf}
\end{figure}

\noindent{\em 2. Adaptive quantization speeds up inference.}
Fig.~\ref{fig:motive_perf} shows how quantization performs with different device types and input shapes. The latency is measured on a single layer of OPT-30b 
with prompt length 512. 
We observe that uniform low-precision quantization may not always result in inference speed-up, due to additional overhead that quantization introduces. 
FP16 precision leads to the fastest inference in many cases. 
If low-precision uniform quantization does not fully occupy the GPU memory, swapping certain layers with faster higher-precision kernels can accelerate the inference process. For instance, when there is remaining memory after uniformly quantizing to INT8, utilizing INT8-FP16 mixed-precision can be beneficial. 

\subsection{Challenges}

\begin{table}[!t]
    \centering
    \caption{Model performance comparison 
    under different layer quantizations.   
    The best results are marked in bold. Unselected layers are retained in FP16.}
    \label{tab:ind_motivation}
\resizebox{0.85\linewidth}{!}{
\begin{tabular}{|c|c|cc|}
    \toprule
    \textbf{Model} & \textbf{Layers Quantized to 4-bit} & \textbf{Avg. Perplexity 
    } & \textbf{Avg. Accuracy (\%) 
    }\\
    \midrule
     \multirow{3}{*}{OPT-1.3b} & 0-8 & \textbf{15.52} & \textbf{62.82} \\
    & 8-16 & 15.78 & 62.49 \\
    & 16-24 & 15.98 & 61.67 \\
   \cline{2-4}
    \multirow{3}{*}{BLOOM-3b} & 0-10 & \textbf{17.65} & \textbf{60.71} \\
    & 10-20 & 17.88 & 60.24 \\
    & 20-30 & 17.94 & 60.37 \\
    \bottomrule
\end{tabular}
}
\vspace{-4mm}

\end{table}

Adopting adaptive mixed-precisions in conjunction with a heterogeneous pipeline model serving poses new challenges. Quantization bit (precision) selection must be considered jointly with layer partition, as the same quantized kernel can perform differently on different GPUs, as shown in Fig.~\ref{fig:two_stage_cost} and Fig.~\ref{fig:motive_perf}. For example, T4 supports fast INT8 due to its tensor core, making the execution time of the 8-bit layer comparable to FP16, while V100's INT8 implementation always incurs longer latency than FP16. Other factors such as micro-batch size, prompt length, and token generation number also affect the kernel speed and pipeline bubble in prefill and decode phases. 
To produce an optimized inference execution plan, we should take into account all these factors,
which results in a complex problem with a very large solution space. 

{\em First}, determining the optimal inference execution plan requires an accurate estimation of memory and latency across devices under different precisions. Profiling every possible combination of precision, GPU type, and input shape for all partition cases would be very time-consuming. 
An efficient cost model is needed to reduce the overhead.
{\em Second}, different layers in an LLM 
 may exhibit different 
 sensitivities to quantization, in terms of model performance impact, when quantized to the same bit. Table~\ref{tab:ind_motivation} shows that selecting different layers of LLMs for quantization can render different model qualities.
  This finding highlights the importance of identifying a suitable layer quantization sensitivity indicator to guide bits selection, achieving the goal of reducing memory waste and promoting model quality simultaneously. 
{\em Last}, due to the 
large solution space of our joint decision-making problem, offline search for optimal solutions can still be time-consuming. 
An efficient algorithm is in need to effectively prune the solution space.


We design \sysplain{} to handle all these challenges and achieve significant performance gains of LLM serving on heterogeneous clusters.
\section{\sysplain{} Overview}\label{sec:sys_overview}

\begin{figure}[t]
    \centering
    \includegraphics[width=0.8\linewidth]{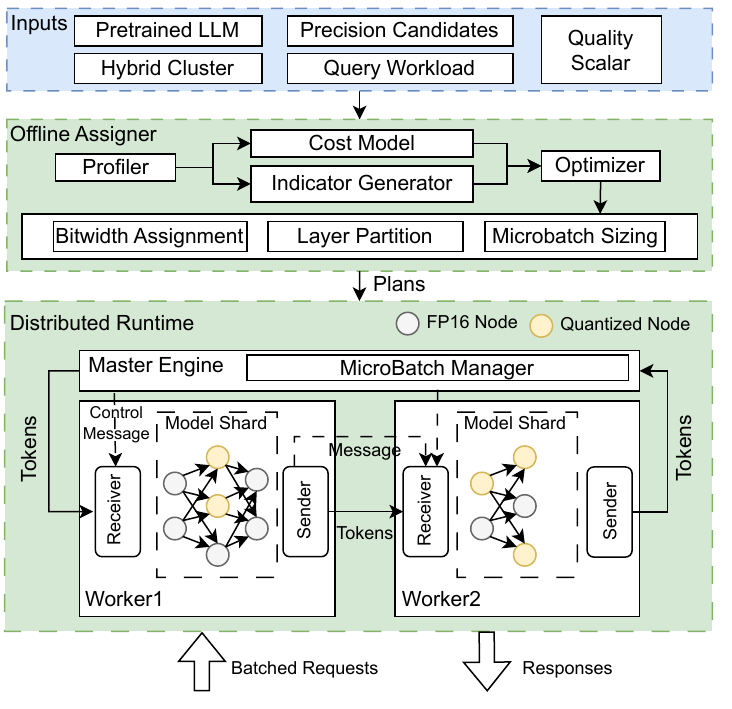}
    \caption{\sysplain{} Overview}
    \label{fig:sys_overview}
\end{figure}


\sysplain{} includes an offline assigner and a distributed model inference runtime. A system overview is given in Fig.~\ref{fig:sys_overview}. 

The offline assigner makes optimized decisions on model layer partition, micro-batch sizing, and quantization bit assignment to each layer. It collects user inputs including the pre-trained LLM, devices 
and their resource configurations in the heterogeneous cluster, precision candidates, query workload characteristics (the prompt length, token generation length, and batch size), and a  `quality scalar' that represents 
user's level of concern for 
mode quality (Sec.~\ref{sec:optimizer}). 
The cost models include: (i) an analytical memory model which takes model meta-information such as hidden space size and decoder layer number as input and predicts the GPU memory occupation for a model shard with its 
mixed-precision plan; 
(ii) a latency cost model, which predicts the 
execution latency of a model shard based on inference latency samples of a single decoder layer collected by the profiler on different GPUs. The Indicator Generator is responsible for producing an indicator that quantifies the  
model performance perturbation introduced by a 
quantized layer under a specific bit. 
The optimizer derives the bit assignment, layer partition, and micro-batch sizing using the indicator and the cost models. 

The distributed runtime 
executes the plans generated by the assigner and conducts LLM generative inference. The master engine 
handles preprocessing and postprocessing for token generation, such as embedding lookup and process logits into a predicted token,  
and micro-batch sizing for different generation phases. Each worker process is responsible for one pipeline stage and is located on a different GPU.  


\section{Assigner Design}\label{sec:assigner}
\begin{table}[t]
\centering
\caption{Notation}
\begin{small}
    \resizebox{\linewidth}{!}{
    \begin{tabular}{|c|c|c|c|}
    \hline
     $h_1$ &  Hidden dimension of Transformer layers &  $h_2$&  Hidden dimension of 2nd MLP layer\\
    \hline
    $v$&  Batch size & $s$& Prompt length \\
    \hline
    $t$& Index of current generated token & $bit$&  Bitwidth of the current layer \\
    \hline
    $d_t$&  Dimension of word embedding projection & $d_p$&  Dimension of position embedding  \\
    \hline
    $vocab_{s}$&  Vocabulary size & $pos_s$ & Max position embeddings \\
    \hline
    \end{tabular}
    \label{tab:notation}
    }
\end{small}
\end{table}

\subsection{Cost Model}\label{sec:assigner_cost_model}
\textbf{Memory Cost Model.} Memory is a first-class citizen in LLM serving systems. 
The peak memory usage of pipeline LLM serving is 
largely due to the model weights, the KV cache for all requests, and the peak temporary memory required by the model layers. 


\textit{Weight Storage.}
The model weight storage is dominated by embedding weights, projections convert the hidden dimension into the word embedding dimension at the model's head and tail, 
and linear weights inside the decoder layers. The embedding weights consist of (1) token embeddings: $vocab_s \times d_t$ (refer to Table  \ref{tab:notation} for notation); (2) position embeddings: $pos_s \times d_t$; and (3) projections (only present when $h_1 \neq d_t$): $2 \times h_1 \times d_t$. The LM head is a single linear layer with weight shape $vocab_s \times d_t$. Since the embeddings and LM head make up a very small portion of the LLM (e.g., 1.4GB out of a 60GB OPT-30b model), 
LLM quantization \cite{frantar2023gptq, dettmers2022llmint8} typically does not quantize this part of the model, which remains in FP16 format. We adopt the same practice, and the memory requirement for them (in bytes) can be summarized as $(vocab_s \times d_t + d_p \times h_1 + 2 \times h_1 \times d_t + vocab_s \times d_t) \times 2$.

For decoder layers, only linear and layer norm layers contribute to memory consumption. For self-attention, the parameters consist of (1) QKV, OUT: $h_1^2$; (2) Layernorm: $4 h_1$ for normal layernorm and $2 h_1$ for RMSNorm~\cite{rmsnorm}. For FFN, the parameters are: (1) 2 MLP: $h_1 \times h_2$; (2) Layernorm: $2 h_1$. The weight of the linear layer can be quantized, 
making the memory requirement for decoder layers with quantization precision $bit$ is: $(4 \times h_1^2 + 2 \times h_1 \times h_2) \times \frac{4\times bit}{32} + 6 \times h_1$ or $4 \times h_1$.

\textit{KV Storage Modeling.}
Like in other 
frameworks~\cite{FTransformer}, \sysplain{} reserves the KV cache 
with a size of maximum sentence length, combining the maximum prompt length $s$ and token generation number $n=t_{max}$ 
, to ensure that there is enough space for subsequent token generation. 
For batched requests, the memory size (in bytes) required by the KV cache can be estimated as $2\times v(s+n)h_1 \times \frac{4\times bit_{kv}}{32}$, where $bit_{kv}$ is the bitwidth used to represent each element in the KV cache. 

\textit{Peak Temporary Memory.} 
Temporary memory required by operators depends on many factors including precision, kernel implementation, and the cache allocator mechanism of the DNN framework. 
We consider a worst-case scenario in evaluating the peak memory required by all involved operators inside the embedding layer and one decoder layer 
in both prefill and decode phases. 

\noindent
\textbf{Latency Cost Model.} Computation intensity varies 
across the prefill and decode phases. 
For example, NVIDIA V100 GPU has an arithmetic intensity of 139 (125TFLOPS / 900 GB/s); the arithmetic intensity during the decode phase of inference over OPT-175b and 30b models for a batch size of 32 and prompt length of 512 is 
48 and 43, respectively. 
On the other hand, execution of the prefill phase on the models incurs arithmetic intensity of 9553 and 6354, respectively, showing that the prefill phase is more computation intensive.

Therefore, we model the execution time of the prefill phase as a function of FLOPs, based on $v, s, vs$ and $vs^2$. 
The decode phase is dominated by memory access; we hence use the total
number of bytes accessed(also called MOPs), 
to model decoding time, based on parameters $v, v(t+s)$ and $(t+s)$.
We profile the execution time of each phase on one decoder layer 
under different precisions with common prompt lengths and batch sizes. We then use interpolation among the sample points to obtain a linear regression model for the execution time of one decoder layer in each phase. We choose linear regression because, in LLM serving, GEMM takes more than 80\% latency~\cite{du2022energonai} and is either FLOPs and MOPs related, while the other operators scaled with MOPs, thus workload can be shaped and scaled by the previous parameters. 
The latency of a model shard can be obtained by summing up the latencies of all involved decoder layers with respect to their precisions.


\subsection{Indicator of Model Perturbation by Quantization}\label{sec:m_ind}

We build performance indicators for low-precision weight-only kernels. INT8 kernel in this paper incurs little performance dradation~\cite{dettmers2022llmint8}, we take the same indicator format with weight-only kernels for simplicity. State-of-the-art weight-only quantization of LLMs focuses on linear operators and
~\cite{frantar2023gptq, lin2023awq, dettmers2023spqr} 
typically target the following objective:
\begin{equation}
\tiny
    \mathbf{{Q}^{*}} = \operatorname*{arg\,min}_{Q} \mathcal{L}(\tilde{\mathbf{W}}), \quad\quad\quad
    \mathcal{L}(\tilde{\mathbf{W}})=\|\mathbf{W}\mathbf{X} - \tilde{\mathbf{W}}\mathbf{X} \|_2^2
\end{equation}

\noindent Here $\mathcal{L}$ is the loss function, 
typically the minimum square error (MSE). $\mathbf{W}$ denotes the set of original FP16 weights of a decoder layer, and $\tilde{\mathbf{W}}$ is the set of quantized weights by quantization method $Q$, i.e., $\tilde{\mathbf{W}} = Q(\mathbf{W})$. $\mathbf{X}$ is the input feature, which refers to the layer input that corresponds to a small set of data points running through the network~\cite{frantar2023gptq}. 
The goal is to identify the quantization method $Q^*$ which 
minimizes the loss.
Previous research~\cite{dong2019hawq} has 
used the eigenvalues of the Hessian matrix $\mathbf{H}$ of $\mathcal{L}$ with respect to $\mathbf{W}$ 
to measure a layer's sensitivity (error term) to quantization, 
as $\omega = \lambda \|Q(\mathbf{W}) - \mathbf{W}\|_2^2$, where $\lambda$ is the top eigenvalue 
of Hessian $\mathbf{H}$. 
It requires computation of Hessian and quantization error ($\|Q(\mathbf{W}) - \mathbf{W}\|_2^2$) with respect to different precisions, incurring large computation overhead. 

We adopt a different approach to describe a layer's sensitivity upon quantization. One key observation is 
that the 
quantization error originates from the $Round$ function. For a vector $\mathbf{x}$, 
$Round$ rounds each of its elements $x$ to $\floor{x}$ or $\ceil{x}$.
We consider the round variance of quantization 
for two widely applied rounding methods, i.e., deterministic and stochastic~\cite{wan2023adaptive}
, and derive an upper bound of the output variance introduced by quantization. 

\begin{theorem}\label{thm:output_var}
The variance of a linear operator's output after weight-only quantization using stochastic or  deterministic rounding 
is:
\begin{equation}
\tiny
Var[\tilde{\mathbf{W}}\mathbf{X}] = \left\{
\begin{aligned}
& Var[\mathbf{W}\mathbf{X}] + D_\mathbf{W}S_\mathbf{W}^2 \frac{1}{4} Var[\mathbf{X}], &\mbox{\small Deterministic}\\
& Var[\mathbf{W}\mathbf{X}] + D_\mathbf{W}S_\mathbf{W}^2 \frac{1}{6}(\mathbb{E}[\mathbf{X}]^2 + Var[\mathbf{X}]), &\mbox{\small Stochastic}
\end{aligned}
\right.
\end{equation}

\noindent where $D_\mathbf{W}$ is the dimension of model weights $\mathbf{W}$ and $S_\mathbf{W}$ is the scaling factor.
\end{theorem}

\noindent The theorem shows that 
the variance introduced by quantization in each linear operator is 
proportional to the dimension and scaling factor of the model weights. 
The scaling factor $S_\mathbf{W}$ 
is typically defined as $S_\mathbf{W} = \frac{\mathbf{W}{max} - \mathbf{W}{min}}{2^{b} - 1}$ (for asymmetric quantization), 
or $S_\mathbf{W} = \frac{max(abs(\mathbf{W}{max}), abs(\mathbf{W}{min}))}{2^{(b - 1)} - 1}$ (symmetric quantization), where $\mathbf{W}{max}$ and $\mathbf{W}{min}$ are the largest and smallest weight values in $\mathbf{W}$. Given 
$\mathbf{W}$, the scaling factor is a function of quantization bitwidth $b$, denoted as $S_\mathbf{W}(b)$.



\begin{proposition}[Variance Indicator] We measure the quantization sensitivity of a decoder layer $i$ using the estimated quantization variance of the layer's output, i.e., 
\begin{equation}\label{eq:ind}
\tiny
    \omega_{i,b} = \sum_o^{O_i} D_{\mathbf{W}_{o}} (S_{\mathbf{W}_{o}}(b_i))^2 G(\mathbf{X}_{o}) 
\end{equation}
\noindent where $O_i$ is all linear operator within a layer, $W_o$ represents the weight of linear operator $o$, 
$X_o$ is the input feature, 
and $G(\mathbf{X})$ equals $\frac{1}{4} Var[\mathbf{X}]$ for deterministic or $ \frac{1}{6}(\mathbb{E}[\mathbf{X}]^2 + Var[\mathbf{X}])$ for stochastic, respectively. 
\end{proposition}

\noindent The variance indicator $\omega$ models the extra variance of output 
of a layer due to weight quantization. 
We use this indicator to rank the model performance impact of different quantization precisions for different layers.
Operations in $G(\mathbf{X})$, i.e., mean and variance, 
are elementwise, with greatly reduced computation complexity as compared to Hessian calculation 
($\mathcal{O}(D_{\mathbf{W}_{i}}D_{\mathbf{X}_{i}})$ vs.~$\mathcal{O}(D_{\mathbf{W}_{i}}D_{\mathbf{X}_{i}}^2)$). 
The missing proofs can be found in supplementary materials.

\subsection{Optimizer}\label{sec:optimizer}

We present an iterative algorithm (Algorithm~\ref{algo:layer_part_bit}) to decide the quantization bitwidth for each decoder layer, micro-batch sizes, and LLM model partition 
and on each device, to strike the best balance between inference latency and model quality degradation. 
The algorithm explores potential device topology orderings and micro-batch sizes for prefill and decode phases; given a device topology ordering and micro-batch sizes, we solve an integer linear program (ILP) 
to determine the most suitable bitwidth assignment and layer partition among the devices. 

\noindent\textbf{Bidwidth Assignment and Layer Partition.}

We use 
binary variable $z_{i,j,b}$ to denote whether layer $i$ is assigned to device $j$ with quantization bitwidth $b$ (1) or not (0). 
${B, \eta, \xi}$ 
denote 
the global batch size, the micro-batch size in the prefill phase, and the micro-batch size in the decode phase, respectively. $L$ is the number of layers in the LLM, and $n$ is the token generation number. 
We suppose input sequences within a batch are padded to the maximal prompt length $s$.
There are $N$ devices, denoted as $j \in \{1, 2, ..., N\}$. $M_j$ is the memory capacity of device $j$. 
$BITs$ is the set of available bitwidth choices, e.g., $BITs=\{3, 4, 8, 16\}$. 

$T_{max}^{pre}$ and $T_{max}^{dec}$ denote the maximum singe-stage latency among pipeline stages in the prefill and decode phase, respectively. $T_{pre}$ and $T_{dec}$ represent 
execution time of the whole model in the prefill phase and decode phase, respectively.
We aim to minimize both inference latency and model performance variance, taking into account both serving speed and model quality. The user's concern for model quality degradation is weighted through coefficient $\theta >0$, with a smaller $\theta$ trading off more model quality over inference acceleration. 



The first parenthesized term in the objective (\ref{eq:objective}) represents the end-to-end serving latency for a batch's token generation. In a pipeline-parallel serving system, the latency of serving a batch is the execution time of all pipeline stages plus $\mu-1$ times the time taken by the slowest stage, where $\mu$ is the number of micro-batches~\cite{zheng2022alpa}. In our LLM serving system, the end-to-end inference latency consists of the 
execution time of prefill and decode phases, corresponding to micro-batch numbers $\mu_{pre} = \ceil{\frac{B}{\eta}}$ and $\mu_{dec} = \ceil{\frac{B}{\xi}}$ for the two phases, respectively. 
Given 
$n$ tokens to generate, the end-to-end latency is the sum of the prefill time of the first token and the decode time of the remaining ${n} - 1$ tokens. 
The second term in the objective corresponds to overall model quality degradation (measured by our variance indicator). 

$T_{max}^{pre}, T_{max}^{dec}, T_{pre}$ and $T_{dec}$  
are contingent upon $\mathbb{Z}$ (the vector of all decision variables $z_{i,j,b}$) as in constraints (\ref{eq:cons_1})-(\ref{eq:cons_4}) 
, where $T_{pre,j}$ is execution time on device j. $l_{i,j,b}^{s, 0}$ represents the average prefill computation time per-batch under prefill micro-batch size $\eta$, and $l_{i,j,b}^{s, \frac{{n}}{2}}$ is the average decode computation time per-batch under decode micro-batch size $\xi$, where $i,j,b$ refers to the layer index, device index, and bitwidth, $s$ is the prompt length. We half the token number ($\frac{{n}}{2}$) for time estimation since decode cost increases linearly with each additional token in the past sequence for the next token. 
Costs are obtained from the latency cost models in Sec.~\ref{sec:assigner_cost_model}. 
Communication in our system is asynchronous, as specified in constraint (7), $P_{pre}$ and $P_{dec}$ denote the transmission data size in the prefill and decode phases and $f_j$ is the communication bandwidth between device $j$ and its successor. 

\begin{figure}[t]
    \begin{equation}
    \tiny
    \begin{aligned}
                \min_{\mathbb{Z} }\quad & (\ceil{\frac{B}{\eta} - 1} T_{max}^{pre} + \ceil{\frac{B}{\xi} - 1} ({n}-1)T_{max}^{dec} + T_{pre} + T_{dec}) \\ 
    & + \theta \sum_{j=1}^{N} \sum_{i=1}^{L}  \sum_{b \in BITs} z_{i, j, b} \omega_{i,b}  &\label{eq:objective}
    \end{aligned}
    \end{equation}
    \scalebox{0.8}{\parbox{\linewidth}{%
    \small
    \begin{align}
    & T_{max}^{pre} \geq T_{pre,j} = \sum_{i=1}^{L} \sum_{b \in BITs} z_{i,j,b} l_{i,j,b}^{s,0}, \quad \forall j=1,...,N \label{eq:cons_1} \\
            & T_{max}^{dec} \geq  T_{dec,j} = \sum_{i=1}^{L} \sum_{b \in BITs} z_{i,j,b} l_{i,j,b}^{s,\frac{{n}}{2} }, \quad \forall j=1,...,N  \label{eq:cons_2} \\
    & T_{max}^{pre} \geq \frac{P_{pre}}{f_j}, \quad
    T_{max}^{dec} \geq \frac{P_{dec}}{f_j}, \quad \forall j=1,...,N  \label{eq:cons_3} \\
    & T_{pre} = \sum_j^{N} T_{pre,j}, \quad T_{dec} = \sum_j^{N} T_{dec,j}  \label{eq:cons_4}\\
    & \sum_{j=1}^{N} \sum_{b \in BITs} z_{i,j,b} = 1, \quad \forall i=1,...,L  \label{eq:cons_5}\\
    & \sum_{j=1}^{N} z_{i,j,b} = y_{i,b}, \quad \forall i=1,...,L, b \in BITs  \label{eq:cons_6} \\
    & \sum_{b \in BITs} z_{i,j,b} = u_{i,j}, \quad \forall i=1,...,L, j=1,...,N,  \label{eq:cons_7} \\
    & \sum_{i=1}^{L} \sum_{b \in BITs} z_{i,j,b} M_{i,b}^{s+n} \leq M_j, \quad \forall j=2,...,N  \label{eq:cons_8} \\
    & \sum_{i=1}^{L} \sum_{b \in BITs} z_{i,1,b} M_{i,b}^{s+n} + M_{emb} \leq M_1  \label{eq:cons_9}\\ 
    & y_{i,b}, u_{i,j}, z_{i,j,b} \in {0,1}, \quad \forall i=1,...,L, j=1,...,N, b \in BITs  \label{eq:cons_12} \\
    & u_{0,0} = 1, u_{L,N} = 1,  \label{eq:cons_13}\\
    & u_{i,j} + u_{i-1,k} \leq 1,  \forall i=2,...,L, j=1,...,N-1, k=j,...,N-1 \label{eq:cons_14}
    \end{align}
    }}
\end{figure}

Constraints (\ref{eq:cons_5}) - (\ref{eq:cons_7}) ensure that only one bitwidth is assigned to a given layer and each layer can only be placed on a single device.
Constraints (\ref{eq:cons_8})-(\ref{eq:cons_9}) guarantee that memory consumption on each device $j$ does not exceed its available memory capacity $M_j$ (which is typically the GPU memory minus those consumed by cuda context), 
where $M_{i,b}^{s+n}$ denotes memory reservation according to the maximum sequence length, using our memory cost model. 
Constraint (\ref{eq:cons_9}) of the first device in the given device ordering 
accommodates the memory requirement, $M_{emb}$, of embeddings for LLM pre or postprocessing 
as well. Constraints (\ref{eq:cons_13})-(\ref{eq:cons_14}) ensure a continuous layer partition solution, as adjacent layer can be only placed on same or neighboring stage, 
where $u_{i,j}$ indicates whether layer $i$ is placed on device $j$. 



We solve the ILP 
using an off-the-shelf solver GUROBI~\cite{gurobi}. 
\begin{algorithm}[t]
\small
\caption{\small Best Inference Execution Plan}
\begin{algorithmic}[1]
\REQUIRE LLM Model $A$ with $L$ Layers, Cluster $C$, Workload $J$ 
\ENSURE Best plan $plm^{*}$
\STATE $\mathcal{G} \leftarrow GetDeviceOrder(C)$, $\mathcal{S} \leftarrow GetMicroBatches(J, C)$
\STATE $plm^{} \leftarrow \emptyset$
\FOR{$(G_1, G_2,\ldots, G_k) \in \mathcal{G}$}
\FOR{$(S_1, S_2,\ldots, S_k) \in \mathcal{S}$}
\STATE Compute $plm_i = F(G_i, J, S_i.\eta, S_i.\xi)$ by solving Optimization~(\ref{eq:objective})
\IF{$plm_i.obj < plm^{*}.obj$} 
\STATE $plm^{*} \leftarrow plm_i$ 
\ENDIF
\ENDFOR
\ENDFOR
\RETURN $plm^{*}$
\end{algorithmic}\label{algo:layer_part_bit}
\end{algorithm}

\noindent\textbf{Device Topology Ordering and Microbatch Sizing.} 
We enumerate all possible combinations of device topology ordering ($GetDeviceOrder$ in line 1 of Alg.~\ref{algo:layer_part_bit}) and micro-batch sizes ($GetMicroBatches$ in line 1 of Alg.~\ref{algo:layer_part_bit}). The device topology ordering is a sequential order of the devices/pipeline stages (one stage on one device) 
and all candidates $\mathcal{G}$ can be derived by permutating the devices. 
The micro-batch size set $\mathcal{S}$ includes sizes $\mu \in [1, B]$. 
Given each combination, we solve the ILP to obtain the corresponding best quantization bitwidth and layer partitions. 

\noindent\textbf{Complexity of Algorithm~\ref{algo:layer_part_bit}.} The solution space size of ILP problem (\ref{eq:objective}) 
is ${\frac{L!}{N!(L-N)!}(|Bits|)}^{L}$, as there are $\frac{L!}{N!(L-N)!}$ possible partitions of the layers and $|Bits|$ possible bitwidths for each layer.
The number of 
algorithm iterations is $|\mathcal{G}||\mathcal{S}|$ at most. Alg.~\ref{algo:layer_part_bit}'s search space is hence $|\mathcal{G}||\mathcal{S}|{\frac{L!}{N!(L-N)!}(|Bits|)}^{L}$. 
This may raise concerns for \textbf{scalability}. We propose several practical optimizations to expedite it.


\textit{Optimization\ \#1: Pruning.}\label{sec:prune}
As discussed in Sec.~\ref{sec:assigner_cost_model}, prefill phase is compute-bound, while the decode phase is memory-bound. GPUs have higher computation capacity than memory bandwidth. Increasing the micro-batch size during the decode phase improves efficiency, but excessively large sizes waste computation capabilities. Evenly partitioning the global batch size across pipeline stages optimizes performance. In the prefill phase, a smaller batch size reduces pipeline bubbles, but extremely small sizes are inefficient. Thus, we enumerate prefill micro-batch size within $[1, \xi]$. 

\textit{Optimization\ \#2: Grouping.}
Grouping multiple layers together and deciding group placement and bitwidth selection can reduce the solution space exponentially. 
For models with a parameter size smaller than 30b, layer grouping is not necessary as $L$ is small. 
For models larger than 30b, grouping layers in sets of 2 is typically sufficient. 

\textit{Optimization\ \#3: Heuristic to solve ILP (\ref{eq:objective}).}

\begin{algorithm}[t]
\small
\caption{\small Bitwidth Transfer. Replacing ILP in Algo.~\ref{algo:layer_part_bit}}
\begin{algorithmic}[1]
\REQUIRE $G_i$, J, $\eta, \xi, BITs$ 
\ENSURE Best plan $plm_{i}^{*}$
\STATE $Adabits = RemoveConstraint(ILP, lat)$ 
\STATE ${obj}_0, plm_{i}^{*}=Adabits(\cdot)$  
\STATE $\mathcal{C}=GetC(BITs)$, $\mathcal{K}=GetK(plm_{i}^{*})$
\WHILE{True}
    \STATE $\mathcal{\grave{K}} = sort(\mathcal{K}), st = \mathcal{\grave{K}}[-1], sol=\emptyset{}$
    \FOR{${pi} \in \mathcal{\grave{K}}[:-2]$}
        \FOR{$c = (b_{st}, b_{pi}, nums) \in \mathcal{C}$}
            \IF{$valid(c, st, {pi})$}
                \STATE Find optimal layers $\{l\}_{pi}, \{l\}_{st}$ 
                \STATE ${obj}^{*} = ExchangePairs(\{l\}_{pi}, \{l\}_{st})$
                \IF{${obj}^{*} > {obj}_0$}
                    \STATE ${obj}_0$ = ${obj}^{*}$, $sol=(c, \{l\}_{pi}, \{l\}_{st})$
                \ENDIF
            \ENDIF
        \ENDFOR
    \ENDFOR
    \IF{$sol \neq \emptyset{}$}
        \STATE $\mathcal{K} = Update(\mathcal{K}, sol)$
        \STATE $plm_{i}^{*} = Update(plm_{i}^{*}, sol)$
    \ELSE
        \STATE break \COMMENT{no valid transformation found}
    \ENDIF
\ENDWHILE
\RETURN $plm_{i}^{*}$
\end{algorithmic}\label{algo:llmpq_heuristic}
\end{algorithm}

GPUs exhibit varying computation capacities, leading to different execution performances for layers while their memory occupation remains fixed. This characteristic allows for precision conversion and layer partition alteration between stages according to transformation rules $\mathcal{C}$. These rules are defined by a three-element tuple $(b_{st}, b_{pi}, nums)$. For example, (4, 8, 2) facilitate the replacement of one 8-bit layer from the pioneer with 2 * 4-bit layers from the straggler. Such transformations increase precision or reduce layer count to accelerate the slowest stage. Leveraging this observation, we propose a heuristic approach called \textbf{bitwidth transfer}, detailed in Algorithm~\ref{algo:llmpq_heuristic}, for solving the ILP problem~(\ref{eq:objective}). Initially, we remove the latency objective from the ILP and solve it under reduced constraints, noted as \textit{adabits}(comparison in Sec.~\ref{sec:adabit_comp})(lines 1-3). We generate potential transformations (line 3), identify the slowest (straggler) and other stages (line 5), and apply possible transformations to improve the target objective value iteratively (lines 6-16). The heuristic is effective in most cases, particularly when KV size does not dominate memory occupation. Further discussion on its usage is provided in Sec.~\ref{sec:expedit}. 

\section{Implementation}
We have implemented \sysplain{} using PyTorch-2.0.0~\cite{Paszke_PyTorch_An_Imperative_2019} with over 6000 LoCs (1355 LoCs for Assigner). 
We extend models on HuggingFace~\cite{HF_transformers} (transformers-4.28.0) to support pre-allocated KV cache 
and adaptive quantization. 
We implement pipeline serving and a thread-safe micro-batch manager on top of the heterogeneous pipeline in ~\cite{hu2021pipeline} with asynchronous communication among stages. 


\noindent
\textbf{On-The-Fly Quantizer}
To optimize the utilization of low-caliber GPUs with smaller DRAM that may frequently experience precision changes, we have developed a specialized and efficient plugin for on-the-fly quantized model loading. In this approach, we have decoupled the integrated model weight into module-level weights. During runtime, we determine the granularity of processed weights by overlapping the disk-to-CPU weight loading time with the on-GPU model quantization and CPU-to-GPU memory copy. This results in a significant reduction in DRAM required for model loading but also improves recovery speed from the possible failure.  

\noindent
\textbf{API and Commands.} \sys{} provides an entry file for the plan generation for different heterogeneous devices. 
\begin{lstlisting}
llmpq-algo \
 --model-name ${model_name} --model_size ${model_size} \ 
 --device_names "${device_names[@]}" \ 
 --device_numbers "${device_numbers[@]}" \ 
 --omega_file $omega_file \ # indicator file
 --global_bz $batch_size --s $s --n $n \ # workload
 --theta $theta \ # user scalar
 --<group $group_size> <--shaq-efficient> \ # faster
 --<fit/use_profiler_prediction>  # use cost model or profiled result
\end{lstlisting}
The output strategy can be launched directly. If the same GPU type is located on the same node, other configurations, such as ranks, will be derived automatically and registered to the distributed runtime. Alternatively, \textbf{distributedconfigs} same as those in PyTorch can be used to launch the strategy, but the \textbf{noauto} flag must be specified.
\begin{lstlisting}[language=bash]
llmpq-dist --strat_file_name $strategy_file_path \
<--master_addr --master_port>/<distribtedconfigs --no_auto> 
\end{lstlisting}

\section{Evaluation}
\subsection{Experimental Setup}

\textbf{Models \& Precisions.} We 
run BLOOM~\cite{scao2022bloom} and OPT~\cite{Zhang2022OPTOP} model families, focusing on 
middle- and large-sized models, specifically OPT-13b, 30b, 66b, and BLOOM-176b. 
We evaluate candidate precisions: $BITs=\{3,4,8,16\}$. 


\noindent\textbf{Baselines.}
We compare \sys{} with three baselines: (1) \textit{PipeEdge}, where we apply uniform quantization and use PipeEdge~\cite{hu2021pipeline} for heterogeneous layer partition. (2) \textit{Uniform}, which uses uniform quantization, evenly partitions the model layers among devices and decides micro-batch sizes that minimize the 
inference latency, mimicking the policy of existing serving systems such as HF-Transformers~\cite{HF_transformers} 
and Deepspeed~\cite{ Aminabadi2022DeepSpeedIE}. (3) \textit{Offloading}, where we adopt CPU and disk swapping in FlexGen~\cite{flexgen} to maximize the throughput of token generation for low-calibre GPUs, we adopt even partition for this method. 
For (1)(2), we keep lowering the quantization bitwidth from the maximum (i.e., FP16) until the model can fit into the devices or no feasible solutions are available. For (1)(3), we use the same micro-batch size for prefill and decode phases by partitioning the global batch size by the number of pipeline stages. 
\textit{FlexGen} is specialized for OPT models and thus has \textit{no results on BLOOM models}. We did not conduct a comparison with vLLM~\cite{vllm} as it primarily focuses on the online task, and the paged attention mechanism is of no use when dealing with fixed token generation numbers. Also, vLLM didn't support pipeline parallelism, making the comparison unfair in our case. 

\noindent\textbf{Metrics.} We evaluate LLM serving performance by (1) token generation throughput, (2) end-to-end serving latency of one batch, and (3) model quality, using perplexity (PPL) on WikiText2~\cite{wikitext2}, Penn Treebank (PTB)~\cite{ptb} and C4~\cite{c4}. The weight calibration data 
consists of 128 randomly selected 2048-token segments from the C4 dataset~\cite{c4}. 

\noindent\textbf{Workload.} 
We use synthetic datasets 
following the 
prompt length setup in the DeepSpeed paper ~\cite{Aminabadi2022DeepSpeedIE}, i.e., 128 and 512. By default, we pad input prompts to 512 tokens, use an input batch size of 32, 
and set the number of tokens to be generated to $n=100$. We follow the same setup as in ORCA~\cite{orca} 
to never emit the EOS but continue to generate tokens until reaching the expected token generation length. 


\noindent\textbf{Heterogeneous Clusters.}
Devices/nodes are in our production cluster. We construct a number of heterogeneous clusters for model serving (clusters 1-8 in Table~\ref{tab:cluster_configs}), with a mix of common types of GPUs. GPUs of the same type are located on the same node, intra-connected with NV-LINK; Clusters 1,2,9,10,11 are on a single node and others consist of two nodes. Nodes in Clusters 3,5,8,11 are interconnected with 800Gbps Ethernet; 4,6, and 7 with 100Gbps Ethernet. All GPUs are equipped with GB/s SSD; Each node is equipped with two CPUs, P100 nodes with Intel Xeon CPU E5-2630 v4 2.2GHz, 64G RAM, V100 and A800 with Intel Xeon Gold 6230 2.1GHz, 128G RAM and 450G RAM, T4 with Intel Xeon Platinum 8260 CPU, 108G RAM, A100-40G with AMD EPYC 7H12 64-Core, 256G RAM. OS: Ubuntu 20.04.6 LTS. We also show \sys{}'s performance on several homogenous clusters (clusters 9-11 in Table~\ref{tab:cluster_configs}). 

\noindent\textbf{Experiment Settings} $\theta$ is handtuned in main experiments. Table~\ref{tab:main},~\ref{tab:homo} has an average solving time 18.38s using GUROBI (max: 115.981s). Detailed $\theta$, solver setup, and overhead table are provided in Appendix~\ref{sec:app_exp_setup}. The model size to run on each cluster is decided such that the total weight size of the 
non-quantized model is comparable to the overall device memory capacity in the cluster.

\begin{table}[t]
\centering
\caption{Cluster Configurations}
\resizebox{\linewidth}{!}{
\begin{tabular}{|l|l|l|l|l|l|}
\hline
\textbf{Cluster} & \textbf{Devices} & \textbf{Model Size} & \textbf{Cluster} & \textbf{Devices} & \textbf{Model Size}  \\ \hline
 1 & 1xV100-32G & 13b &  2 & 1xA100-40G & 13b \\ \hline
 3 & 3xT4-16G + 1xV100-32G & 30b  &  4 & 3xP100-12G + 1xV100-32G & 30b \\ \hline
 5 & 4xT4-16G + 2xV100-32G & 66b  &  6 & 2xV100-32G + 2xA100-40G & 66b  \\\hline 
 7 & 4xV100-32G + 4xA100-40G & 176b &  8 & 4xV100-32G + 2xA800-80G & 176b \\ \hline
 9 & 4xT4-16G & 30b &  10 & 4xV100-32G & 66b   \\ \hline
 11 & 4xA800-80G & 176b  & & &\\ \hline
\end{tabular}\label{tab:cluster_configs} 
}
\end{table}

\subsection{Fidelity of Cost Models}
\begin{figure}[t]
    \centering
    \includegraphics[width=\linewidth]{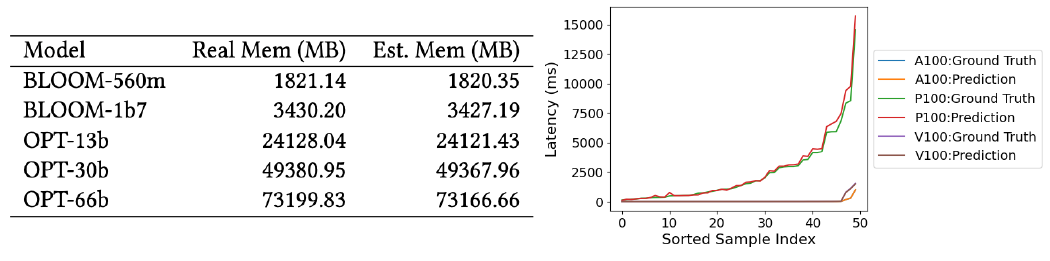}
    \caption{Comparison of memory and latency reported by the cost models and obtained in real systems.}
    \label{fig:CostModel}
\end{figure}

We evaluate our memory cost model on 
BLOOM of sizes 560m and 1b7, and OPT of 13b, 30b, and 66b, with prompt length uniformly sampled between 128 and 512, the batch size chosen among 2, 4, and 8, generated token length sampled between 100 and 200, and randomly generated precision setting from the available bitwidth set. 
We consider the memory consumption of model weights and KV caching here and compare the predicted memory usage with those collected from real systems. 
We also create 
50 unseen workloads 
with different precisions, batch sizes (3,5 or 7), prompt lengths, and past sequence lengths 
(384 or 768) for each device, 
evaluate our latency cost model on them. 
Fig.~\ref{fig:CostModel} shows that the error of the memory cost model is almost negligible, 
and the average error of the latency cost model is 
less than 6\%. 

We observed that, during the prefill phase, the cost of observations typically increases linearly with the workload. However, it is noteworthy that in the decode phase, a notable difference in latency occurs only when a substantial change in context length (50-100) is present.

\subsection{Serving in Heterogeneous Clusters}

\begin{table*}[!t]
\caption{Serving performance comparison. 
The best results 
are marked in bold. The missing results are due to OOM. The $\times$ is derived comparing with 
the PipeEdge baseline.} 
\label{tab:main}
\begin{small}
\resizebox{\textwidth}{!}{
\renewcommand{\arraystretch}{1.2}
\begin{tabular}{c|c|ccccc|c|c|ccccc}
    \toprule
    \textbf{Model Size} & \textbf{Cluster} & \textbf{Model} & \textbf{Scheme} & \textbf{PPL} & \textbf{Latency (s)} & \textbf{Throughput (Token/s)} & \textbf{Model Size} & \textbf{Cluster} & \textbf{Model} & \textbf{Scheme} & \textbf{PPL}  & \textbf{Latency (s)} & \textbf{Throughput (Token/s)} \\
    \midrule
    \multirow{10}{*}{13b}& \multirow{5}{*}{1}& \multirow{5}{*}{OPT} & \basef{} & 11.78 & 233.77 & 13.69 & \multirow{10}{*}{66b}& \multirow{5}{*}{5}& \multirow{5}{*}{OPT} & \basef{} & 10.50 & 750.84 & 4.26\\
    & & & $\bases{}^{*}$  & 11.23 & 57.59 & 55.57(4.06$\times$) & & & & \bases{} & $\dagger$ &$\dagger$ &$\dagger$ \\
   & & & \basefr{} & 11.22 & 174.88 & 18.30(1.34$\times$) & & & & \basefr{} & $\dagger$  & $\dagger$ & $\dagger$ \\
  & & & $\baseff{}^{*}$ & 11.23 & 50.20 & \textbf{63.74(4.66$\times$)} & & & & \baseff{} & \textbf{10.34} & 704.93  & 5.11 (1.20$\times$) \\
   & & & $\sysplain{}^{*}$  & 11.23  & 57.59 & 55.57(4.06$\times$) & & & & \sysplain{} & 10.40(-0.10) & 320.84 & \textbf{9.97(2.34$\times$)}\\
   \cline{3-7}
   \cline{9-14}
   & \multirow{5}{*}{2} & \multirow{5}{*}{OPT} & \basef{} & 11.38 & 30.84 & 103.76 & & \multirow{5}{*}{6} & \multirow{5}{*}{OPT} & \basef{} & 10.34 & 115.03 & 27.82\\
   & & & \bases{} & 11.38  & 30.84 & 103.76 & & & & \bases{} & 10.50 & 431.92 & 7.41(0.27$\times$)\\
   & & & \basefr{} & \textbf{11.22} & 71.09 & 45.01(0.43$\times$) & & & & \basefr{} & 10.33  & 279.05 & 11.47(0.41$\times$) \\
   & & & \baseff{} & 11.23 & 31.11 & 102.87(0.99$\times$) & & & & \baseff{} & 10.34 & 202.32 &15.82 (0.57$\times$) \\
   & & & \sysplain{} & 11.23(-0.14)  & 20.63 & \textbf{155.13(1.50$\times$)} & & & & \sysplain{} & \textbf{10.31(-0.03)} & 68.67 & \textbf{46.60(1.68$\times$)}\\
   \cline{2-7}
   \cline{8-14}
    \multirow{10}{*}{30b}& \multirow{5}{*}{3}& \multirow{5}{*}{OPT} & \basef{} & 10.70 & 146.40 & 21.86 & \multirow{10}{*}{176b}& \multirow{5}{*}{7}& \multirow{5}{*}{BLOOM} & \basef{} & 10.97  & 729.91 & 4.38\\
    & & & \bases{} & 10.78 & 948.90 & 3.37(0.15$\times$) & & & & \bases{} & $\dagger$ & $\dagger$ &$\dagger$ \\
  & & & \basefr{} & 10.70 & 820.72 & 3.90(0.18$\times$) & & & & \basefr{} & $\dagger$  & $\dagger$ & $\dagger$ \\
   & & & \baseff{} & 10.70 & 309.95 & 10.32(0.47$\times$) & & & & \baseff{} & $\dagger$  & $\dagger$ & $\dagger$ \\
   & & & \sysplain{} & \textbf{10.70}  & 80.60 & \textbf{39.70(1.82$\times$)} & & & & \sysplain{} & \textbf{10.90(-0.07)}  & 427.76 & \textbf{7.48(1.71$\times$)}\\
   \cline{3-7}
   \cline{9-14}
   & \multirow{5}{*}{4} & \multirow{5}{*}{OPT} & \basef{} & 10.78 & 449.55 & 7.12 & & \multirow{5}{*}{8} & \multirow{5}{*}{BLOOM} & \basef{} & 10.97 & 848.98 & 3.77\\
   & & & \bases{} & $\dagger$ & $\dagger$ & $\dagger$ & & & & \bases{} &  $\dagger$ &$\dagger$ &$\dagger$\\
  & & & \basefr{} & 10.70 & 1,348.16 & 2.37(0.33$\times$) & & & & \basefr{} & $\dagger$  & $\dagger$ & $\dagger$ \\
   & & & \baseff{} & 10.70 & 448.18 & 7.14(1$\times$) & & & & \baseff{} & $\dagger$  & $\dagger$ & $\dagger$ \\
   & & & \sysplain{} & \textbf{10.70(-0.08)}  & 214.19 & \textbf{14.94(2.10$\times$)}   & & & & \sysplain{} & \textbf{10.90(-0.07)} &\textbf{294.68} & \textbf{10.86 (2.88$\times$)}\\
   \bottomrule
\end{tabular}
}
\end{small}
\end{table*}
Table~\ref{tab:main} demonstrates that \sys{} achieves the highest inference throughput by dividing the total number of generated tokens in a batch by the corresponding end-to-end latency. 
and the best model accuracy in clusters 3, 4, 6, 7, and 8. In cluster 2, \sys{} incur a negligible perplexity drop (0.01) but achieves a much faster inference speed (1.5$\times$). 
In cluster 6, the perplexity of \sys{} is even better than in the FP16 case. As compared with \basef{} and \bases{}, \sys{} can better utilize memory in heterogeneous devices and conduct phase-aware and precision-aware model partitions. \sys{} also outperforms \basefr{} and \baseff{} in most cases as they suffer from heavy swapping overhead. 
The results on cluster 1 reveal that our micro-batch sizing 
reducing the peak temporary memory needed by the model, allowing the int8 quantized model to fit nicely into the device memory.  

\subsection{Serving in Homogeneous Clusters}\label{sec:eval_serv_heter}

\begin{table}[!t]
    \centering
    \caption{Serving performance comparison 
    in homogenous clusters. The best inference throughput is marked in bold.}
    \label{tab:homo}
    \begin{small}
        \resizebox{\linewidth}{!}{
\begin{tabular}{c|c|cccc}
    \toprule
    \textbf{Model 
    } & \textbf{Cluster} & \textbf{Scheme} & \textbf{PPL} & \textbf{Latency (s)} & \textbf{Throughput (Token/s)}\\
    \midrule
    \multirow{5}{*}{OPT-30b}& \multirow{5}{*}{9}  & \basef{} & 10.78 & 1,045.93 & 3.06\\
    & & \bases{} & 10.78 & 1,045.93 & 3.06 \\
    & & \basefr{} & 10.70 & 1,033.39  & 3.10(1.01$\times$)  \\
    & & \baseff{} & \textbf{10.70}  & 313.46  & \textbf{10.21(3.34$\times$)}  \\
    & & \sysplain{} & 10.75 & 407.75 & 7.85(2.57$\times$)\\
    \midrule
    \multirow{5}{*}{OPT-66b}& \multirow{5}{*}{10}  & \basef{} & 10.33 & 182.47 & 17.54 \\
    & & \bases{} & 10.50 & 477.52 & 6.70(0.38$\times$) \\
    & & \basefr{} & 10.33 & 433.99  & 7.37(0.42$\times$)  \\
    & & \baseff{} & 10.34 & 206.93  & 15.46(0.88$\times$)  \\
    & & \sysplain{} & \textbf{10.33}  & 178.11 &  \textbf{17.97(1.02$\times$)}\\
    \midrule
    \multirow{3}{*}{BLOOM-176b}& \multirow{3}{*}{11} & \basef{} & 10.90 & 49.12 & 65.14 \\
    & & \bases{} & 10.97 & 895.45 & 3.57(0.05$\times$) \\
    & & \sysplain{} & \textbf{10.90}  & 45.45 & \textbf{70.41(1.08$\times$)} \\ 
    \bottomrule
\end{tabular}
}
\end{small}
\end{table}
On homogeneous clusters, 9, 10, and 11, Table~\ref{tab:homo} shows that \sys{} still achieves throughput gains, though 
smaller than on heterogeneous clusters. 
In the case of cluster 9, the performance and perplexity of \sys{} are inferior to that of \baseff{}. This discrepancy is attributed to the limited GPU memory compared to the workload requirement, resulting in high compression and usage of more low-precision kernels. Consequently, the computational speed is slower, but the efficiency of swapping is enhanced. 
Among other cases, \sys{} performs the best on model quality and serving throughput. 

\subsection{Effectiveness of Variance Indicator}

 
To further validate the effectiveness of our model variance indicator, we compare it with random assignment, where $\omega_{i,b}$ is assigned a value sampled from a uniform distribution. In the random indicator, we force higher bitwidth indicator values to be kept smaller than lower bitwidth indicator values within a layer. We also compare our indicator with Hessian-based as discussed in Section~\ref{sec:m_ind}. 
We replace the indicator used in \sys{} 
and adjust $\theta$ in (\ref{eq:objective}) to ensure that different indicators lead to similar inference latency, 
eliminating the influence of value range of the indicator.   
In Table~\ref{tab:ind_res}, we observe that \sys{} 
achieve better perplexity than FP16 on cluster 6. 
On cluster 9, with heavier quantization as mentioned above, 
Hessian-based and our indicators yield the same perplexity, outperforming the pure random indicator. 

\begin{table}[!t]
    \centering
    \caption{Effectiveness of \sys{}'s variance indicator. PPL is compared with Random, while $\times$ is compared with Hessian.} 
    \label{tab:ind_res}
    \begin{small}
        \resizebox{\linewidth}{!}{
\begin{tabular}{c|c|ccc}
    \toprule
    \textbf{Model} & \textbf{Cluster} & \textbf{Method} & \textbf{PPL} & \textbf{Overhead (s)}\\
    \midrule
    \multirow{3}{*}{OPT-66b}& \multirow{3}{*}{6} & Random & 10.33 & 0  \\
    & & Hessian &  10.33 & 25625.44 \\
    & & \sys{} & \textbf{10.31(-0.02)} &  \textbf{434.78(58.15$\times$)} \\
    \midrule
    \multirow{3}{*}{OPT-30b}& \multirow{3}{*}{9}  & Random & 11.04 & 0 \\
    & & Hessian & 10.75 & 15670.87\\
    & & \sys{} & \textbf{10.75(-0.29)} & \textbf{215.60(72.69$\times$)}  \\
    \bottomrule
\end{tabular}
}
\end{small}
\end{table}

\subsection{Serving with Shorter Prompts
}
We next experiment with input prompt length of 128 and 
maximal token generatoin number $n=200$. 
Table~\ref{tab:lat_aware} shows that \sys{} 
achieves substantial inference speed-ups without any accuracy degradation, and even shows accuracy improvements. 
This confirms the correctness of our two-phase latency modeling in \sys{}. We note that the throughput gain of \sys{} in cluster 4 is much lower than that with prompt length 512, which we attribute to the reduced KV cache memory and the fact that smaller prompts and larger token generation numbers make the inference system more akin to the one-phase system that PipeEdge focuses on.


\begin{table}[!t]
    \centering
    \caption{Serving performance comparison 
    under shorter prompts. 
    The best results 
    are marked in bold.}
    \label{tab:lat_aware}
    \begin{small}
        \resizebox{\linewidth}{!}{
\begin{tabular}{c|c|cccc}
    \toprule
    \textbf{Model} & \textbf{Cluster} & \textbf{Scheme} & \textbf{PPL} & \textbf{Latency(s)} & \textbf{Throughput (Token/s)}\\
    \midrule
    \multirow{5}{*}{OPT-13b}& \multirow{5}{*}{1}  & \basef{} & 11.23 & 84.80 & 75.47\\
    & & \bases{} & 11.23 & 84.80 & 75.47(1.00$\times$) \\
    & & \basefr{} & 11.22 & 119.24 & 53.68(0.71$\times$) \\
    & & \baseff{} & 11.23 & 80.35 & 79.65(1.06$\times$) \\
    & & \sysplain{} & \textbf{11.23} & 47.63 & \textbf{134.38(1.78$\times$)}\\
    \midrule
    \multirow{5}{*}{OPT-30b}& \multirow{5}{*}{4}  & \basef{} & 10.70 & 366.54 & 17.46 \\
    & & \bases{} & 10.80 & 281.83 & 22.71(1.30$\times$) \\
    & & \basefr{} & 10.70 & 2,147.03 & 2.98(0.17$\times$) \\
    & & \baseff{} & 10.70 & 681.78 & 9.39(0.54$\times$) \\
    & & \sysplain{} & \textbf{10.70} & 262.34 & \textbf{24.40(1.40$\times$)}\\
    \midrule
    \multirow{5}{*}{OPT-66b}& \multirow{5}{*}{6} & \basef{}& 10.33 & 132.34 & 48.36 \\
    & & \bases{} & 10.33 & 298.99 & 21.41(0.44$\times$) \\
    & & \basefr{} & 10.33 & 408.19 & 15.68(0.32$\times$) \\
    & & \baseff{} & 10.34 & 376.69 & 16.99(0.35$\times$) \\
    & & \sysplain{} & \textbf{10.30(-0.03)}  & 75.98 & \textbf{84.23(1.74$\times$)} \\ 
    \bottomrule
\end{tabular}
}
\end{small}
\end{table}

\subsection{
Approaches Expediting Optimizer Algorithm}\label{sec:expedit}
\begin{table}[!t]
    \centering
    \caption{Effectiveness of Grouping and Heuristic approaches under time limit. 
    The best results are marked in bold.}
    \label{tb:faster}
    \begin{small}
        \resizebox{\linewidth}{!}{
\begin{tabular}{c|c|ccc}
    \toprule
    \textbf{Model} & \textbf{Cluster} & \textbf{Method} & \textbf{Throughput (token/s)} & \textbf{Overhead (s)}\\
    \midrule
    \multirow{3}{*}{OPT-30b}& \multirow{3}{*}{3}  & Group=2 & \textbf{39.70}  & \textbf{1.07} \\
    & & Group=1 & \textbf{39.70(+0)} & 3.29\\
    & & Heuristic & 35.17 & 5.36 \\
    \midrule
    \multirow{3}{*}{OPT-66b}& \multirow{3}{*}{6}  & Group=2 & 39.56 & \textbf{2.70} \\
    & & Group=1 &  \textbf{44.93(+5.37)} & 19.14\\
    & & Heuristic &  28.45 & 7.70  \\
    \midrule
    \multirow{3}{*}{OPT-30b}& \multirow{3}{*}{4}  & Group=2 & 14.72 & 12.29 \\
    & & Group=1 & 13.93(-0.79) & 204.59\\
    & & Heuristic & \textbf{14.94(+0.22)}  & \textbf{1.99}  \\
    \midrule
    \multirow{3}{*}{OPT-66b}& \multirow{3}{*}{10}  & Group=2 & 16.64 & 59.27 \\
    & & Group=1 & 17.57(+0.93) & 127.28\\
    & & Heuristic &  \textbf{17.97(+1.33)} & \textbf{2.11}  \\
    \bottomrule
\end{tabular}
}
\end{small}
\end{table}

In \sysplain{}, we provide two approaches, layer grouping, and a heuristic, to reduce 
and the complexity of the optimizer's bitwidth selection, model partition, and placement. 
We evaluate the inference throughput and the time required to derive the solution 
when applying three strategies ({\em group = 2}, {\em group = 1}, and {\em heuristic}), 
on clusters 3, 4, 6, and 10. {\em group = 2} means group 2 decoder layers together for decision. 
\textbf{We set a 60-second time limit} for the ILP solver.

{\em Group = 1} covers the entire solution space and typically produces better results compared to {\em group = 2} (on clusters 6 and 10), but it introduces a larger overhead, as shown in Table~\ref{tb:faster}. 
On cluster 4, {\em group = 1} cannot find a good solution within the time limit. On cluster 3, {\em group = 1} and {\em group = 2} produce the same solution. Performance of the heuristic largely depends on the starting point produced by {\em adabits} (start point of optimization \#3 in Sec.~\ref{sec:optimizer}). 
It leads to the best throughput with the smallest overhead in clusters 4 and 10.

We highlight the utilization of heuristics significantly enhances the \textbf{scalability} of \sysplain{} in offline workloads: solving time of a cluster comprising two P100, V100, and A100 GPUs each for OPT66B is reduced to 31s.

\subsection{Parameter Sensitivity}

\begin{figure}[t]
    \centering
    \subfigure[Cluster 9 OPT-30b]{
        \includegraphics[width=0.47\linewidth]{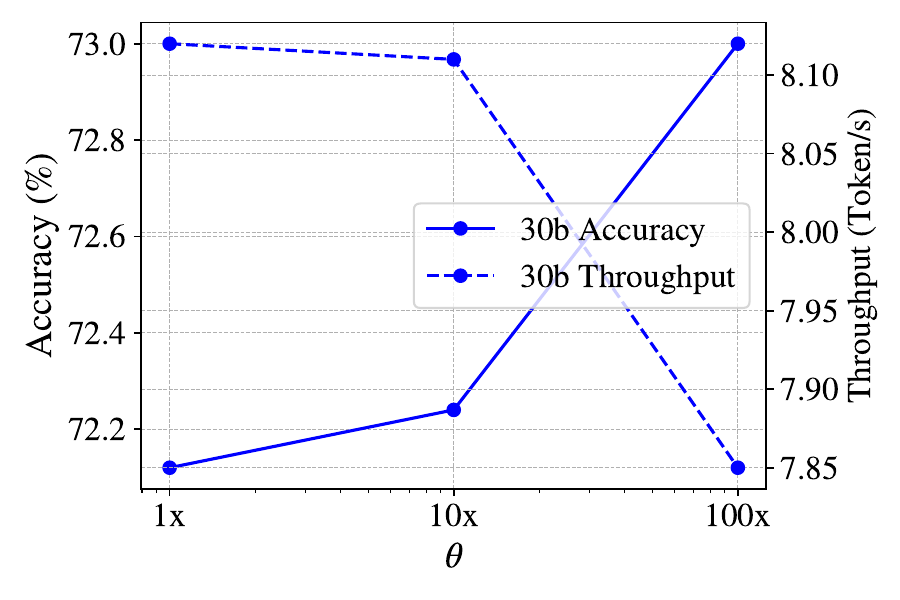}
    }
    \subfigure[Cluster 5 OPT-66b]{
        \includegraphics[width=0.47\linewidth]{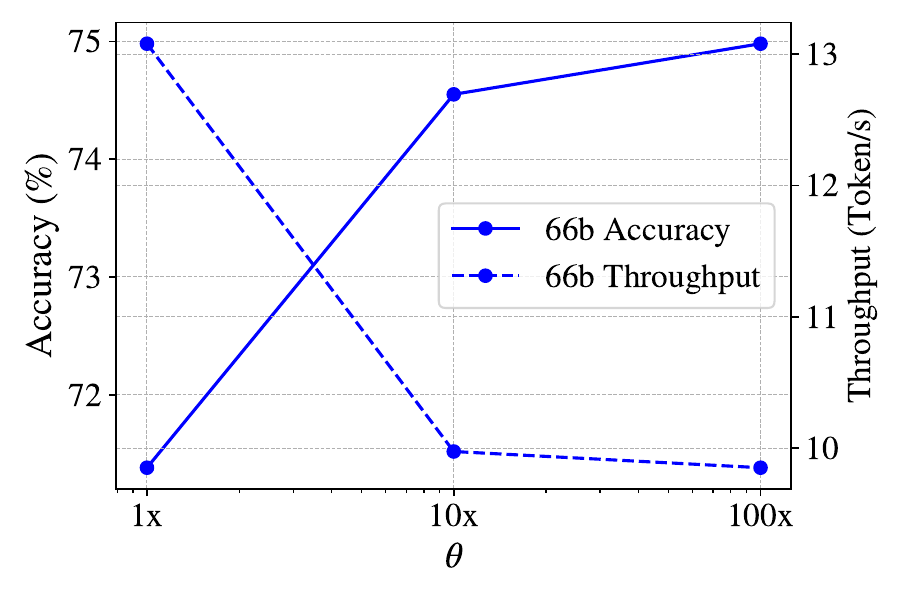}
    }
    \caption{Sensitivity experiments on $\theta$.}
    \label{fig:sens}
\end{figure}

We next investigate the impact of user quality scalar $\theta$ in (\ref{eq:objective}). 
We denote the value of $\theta$ we used in experiment in Sec.~\ref{sec:eval_serv_heter} as $10\times$, scale it by $0.1$ and $10$ to obtain $\theta$ values of $1\times$, $100\times$.
We evaluate model quality and serving throughput of \sysplain{} under each $\theta$ value. 
Fig.~\ref{fig:sens} 
shows that a larger $\theta$ generally results in lower inference throughput and higher model accuracy, as less weight is placed on inference latency and more on model quality in our ILP optimization. 

\subsection{Comparison with Pure Adaptive Quantization}\label{sec:adabit_comp}
\begin{figure}[t]
    \centering
    \includegraphics[width=\linewidth, height=1.0in]{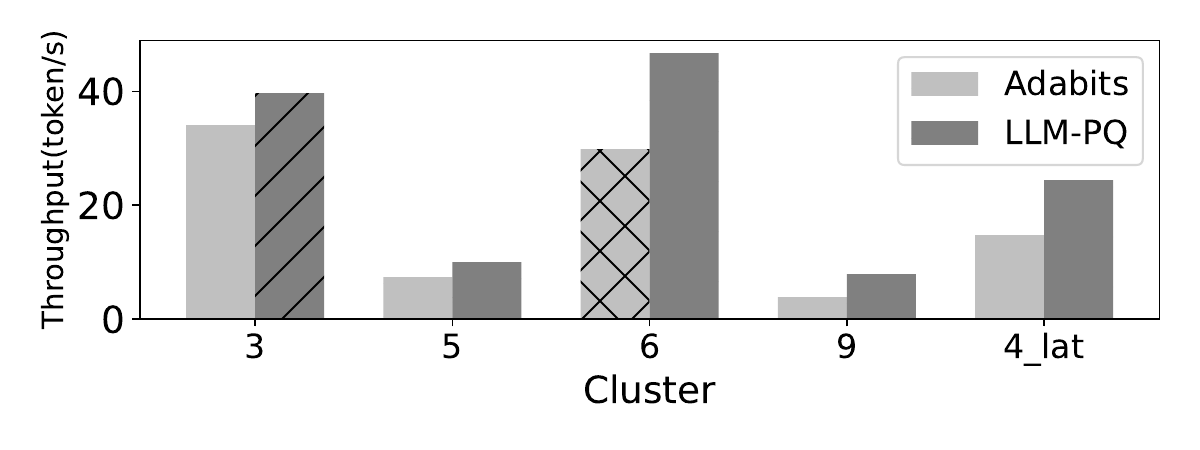}
    \caption{Comparison with pure adaptive quantization.}
    \label{fig:adabits}
\end{figure}

To verify the significance of concurrently considering adaptive bitwidth, layer partitioning, and micro-batch sizing, we further compare \sys{} with {\em adabits} used in the heuristic method. 
We evaluate the performance of {\em adabits} with same model setup 
on clusters 3, 5, and 6, 9 with prompt length 512 and on cluster 4 with prompt length 128. In Fig.~\ref{fig:adabits} 
\sys{} outperforms {\em adabits} in all selected cases.

\section{
Discussions}
\textbf{Search for Tensor Parallelization.}
We did not incorporate tensor parallelism in our serving system implementation due to the favorable characteristics of the pipeline when dealing with heterogeneity, which results in reduced communication requirements. 
It can be readily included in our search space. 
Tensor parallelism heavily relies on the 2-d device mesh configuration, 
and tensor sharding strategies can be searched based on the device mesh enumeration. Given 2 nodes with 8 GPUs per node (totaling 16 devices), we can represent them as a device mesh of size 2×8, 1×16, 4×4, 8×2, or 16×1, where the device communication with different bandwidths for the first and second-dimension, and the tensor-parallel can apply along either the first or second dimension~\cite{zheng2022alpa}. 
As the possible device mesh is limited, it is similar to 
how we enumerate all possible 1-d device orderings. For the above reason, we can view the device along the tensor-parallel dimension as a new device with larger memory and different kernel performance (as tensor-parallel will introduce some communication overhead), and it is still a 1-d partition problem along another axis, which conforms to our solutions.


\noindent
\textbf{Other Quantization Schemes}
There is rapid development in quantization methods for LLM. The latest weight-only quantization methods, such as AWQ~\cite{lin2023awq}, SpQR~\cite{dettmers2023spqr} and QLoRA~\cite{dettmers2023qlora}, AWQ improves kernel efficiency through re-order free quantization and utilizes TensorCore. SpQR improves the accuracy of GPTQ through better outlier detection. QLoRA proposes a memory-efficient 4-bit finetuning method and introduces double quantization to further reduce the memory footprint by quantizing the scalars used in quantization. 
\sys{} views these schemes as candidate quantization schemes, and these new schemes 
can be efficiently integrated into our system.

\noindent
\textbf{Apply to ORCA or vLLM}
ORCA~\cite{orca} introduces iterative-level scheduling, while vLLM~\cite{vllm} possesses an efficient page-attention technology for memory management. \sys{}'s design is orthogonal to both of them. However, unlike the offline task, the online workload is unpredictable, and the available paged memory for Key-Value (KV) storage is affected by quantization level. While the available memory plays a crucial role in influencing throughput when confronted with an infinite number of requests, there is always a trade-off between the speed of quantized operators and the amount of available memory. This trade-off necessitates new design considerations for performance optimization when implementing LLM-PQ at runtime.
\section{Conclusion}
We propose \sysplain{}, an efficient system for LLM serving atop heterogeneous clusters. We 
derive efficient cost models to accurately predict memory occupation and execution latency of mixed-precision LLM serving. We introduce adaptive mixed-precision into the search space of pipeline serving 
and proposed an efficient indicator to guide bitwidth selection in the search process. We jointly consider serving latency in different token generation phases based on various precision settings, micro-batch sizes, and layer partitions, and derive efficient optimized solutions. Our extensive experiments validate the performance of \sysplain{} on a variety of cluster setups, which surpasses state-of-the-art approaches of serving LLM on heterogeneous clusters.

\bibliographystyle{plain}
\clearpage
\bibliography{paper}

\clearpage
\newpage
\appendix
\section{Appendix}

\subsection{Proof of Theorem~\ref{thm:output_var}}
\begin{proof}
Let $\mathbf{X}$ be the input features sampled from a given distribution $D$, the initial variance introduced by weight-only quantization is proportional to the weight dimension and its corresponding scaling factor, and scaled by its input variance. 
The actual scalar multiplication within the matrix multiplication to be $y=\tilde{w}x$, where $\tilde{w} \in Q(\mathbf{W})$ and $x \in \mathbf{X}\sim{D}$. 

In deterministic rounding~\cite{xiao2023smoothquant,frantar2023gptq}, quantized scalar can be either $\hat{w} = \floor{\frac{w-q_w}{s_w}}$ or $\ceil{\frac{w-q_w}{s_w}}$, and the error term $err_w=w-\tilde{w}$ is thus deterministic and bounded by $\pm \frac{1}{2} s_w $. The variance with respect to the output value can be thus formulated as $Var[y] = Var[x(w+err_w)] = Var[y] + \frac{1}{4}s_w^2 Var[x]$, making $Var[\tilde{\mathbf{W}}\mathbf{X}] = Var[\mathbf{Y}] + \frac{1}{4}D_\mathbf{W}S_\mathbf{W}^2 Var[\mathbf{X}]$. 

For stochastic rounding~\cite{chen2021actnn,Liu2022EXACTSG}, scalar $\hat{w} = \floor{\frac{w-q_w}{s_w}}$ with probability $p=\frac{w-q_w}{s_w} - \floor{\frac{w-q_w}{s_w}}$, or up to $\ceil{\frac{w-q_w}{s_w}}$ with probability $1-p$. Suppose $\hat{w} - \floor{\frac{w-q_w}{s_w}} =\sigma \sim Uniform(0,1)$, and $Var[\tilde{w}] = \frac{s_w^2}{6}, Var[\tilde{\mathbf{W}}] = \frac{s_w^2D_{\mathbf{W}}}{6}$, and we always have $\mathbb{E}[\tilde{\mathbf{W}}] = \mathbb{E}[\mathbf{W}]$. $Var[\tilde{\mathbf{W}}\mathbf{X}] = \mathbb{E}[\tilde{\mathbf{W}}]^2Var[\mathbf{X}] + \mathbb{E}[\mathbf{X}]^2 Var[\tilde{\mathbf{W}}] + Var[\tilde{\mathbf{W}}]Var[\mathbf{X}] = \|\mathbf{W}\|^2Var[\mathbf{X}] + \frac{D_\mathbf{W}S_\mathbf{W}^2}{6}(\mathbb{E}[\mathbf{X}]^2 + Var[\mathbf{X}])$
\end{proof}

\subsection{Experiment}\label{sec:app_exp_setup}

\subsubsection{$\theta$ and Solver Setup}
\begin{table}[t]
\centering
\caption{Solver setups for Table~\ref{tab:main} and ~\ref{tab:homo}}
\begin{tabular}{|c|c|c|c|}
\hline
\textbf{Cluster} & \textbf{Group} & \textbf{Heuristic?} & \textbf{$\theta$} \\
\hline
1 & 1 & N & 1 \\
2 & 1 & N & 1 \\
3 & 1 & N & 1 \\
4 & - & Y & 1000 \\
5 & - & Y & 50 \\
6 & 1 & N & 100 \\
7 & 1 & N & 10 \\
8 & 1 & N & 10 \\
9 & 1 & N & 1 \\
10 & - & Y & 1 \\
11 & - & Y & 10 \\
\hline
\end{tabular}
\label{tab:table_solver_setup}
\end{table}

Table~\ref{tab:table_solver_setup} provides the $\theta$ and solver configurations used in both hetero- and homogeneous results for \sys{}.

\subsubsection{Overhead Table}\label{sec:app_overhead}

\begin{table}[t]
\centering
\caption{Problem solving overhead for Table~\ref{tab:main} and ~\ref{tab:homo}}
\begin{tabular}{|c|c|}
\hline
\textbf{Cluster} & \textbf{Overhead(s)} \\
\hline
1 & 0.2977 \\
2 & 0.2977 \\
3 & 2.78127 \\
4 & 2.28628 \\
5 & 9.9239153 \\
6 & 115.981 \\
7 & 44.3031 \\
8 & 19.31674 \\
9 & 1.15838 \\
10 & 2.45544 \\
11 & 3.4 \\
\textbf{AVG} & 18.38195685 \\
\textbf{SLOWEST} & 115.981 \\
\hline
\end{tabular}\label{tab:table_4_5_overhead}

\end{table}

Table~\ref{tab:table_4_5_overhead} presents the solving latency of both hetero- and homogeneous results for \sys{}. We also provide a data point for the three-nodes cluster: Cluster of P100, V100, and A100 GPUs (two each type): solving time with 31s for OPT66B using heuristic.

\end{document}